\documentclass[10pt,twocolumn,letterpaper]{article}

\usepackage{cvpr}              

\usepackage{graphicx}
\usepackage{amsmath}
\usepackage{amssymb}
\usepackage{booktabs}
\usepackage{relsize}
\usepackage{float}

\usepackage[pagebackref,breaklinks,colorlinks]{hyperref}

\usepackage[capitalize]{cleveref}
\crefname{section}{Sec.}{Secs.}
\Crefname{section}{Section}{Sections}
\Crefname{table}{Table}{Tables}
\crefname{table}{Tab.}{Tabs.}

\pdfoutput=1

\def\cvprPaperID{205} 
\def\confName{CVPR}
\def\confYear{2024}

\newcommand{\preddy}[1]{{\color{red}{\textbf{preddy: #1}}}}

\newcommand{\ili}[1]{{\color{blue}{\textbf{ili: #1}}}}

\begin{document}

\title{G3DR: Generative 3D Reconstruction in ImageNet}


\author{Pradyumna Reddy\textbf{\textsuperscript{*}} 
\quad
Ismail Elezi\textbf{\textsuperscript{*}}
\quad
Jiankang Deng
\\
Huawei Noah's Ark Lab UK\hspace{1cm} \\
}


\maketitle
%

\begin{abstract}
We introduce a novel 3D generative method, Generative 3D Reconstruction (G3DR) in ImageNet, capable of generating diverse and high-quality 3D objects from single images,
addressing the limitations of existing methods. 
At the heart of our framework is a novel depth regularization technique that enables the generation of scenes with high-geometric fidelity. 
G3DR also leverages a pre-trained language-vision model, such as CLIP, to enable reconstruction in novel views and improve the visual realism of generations. 
Additionally, G3DR designs a simple but effective sampling procedure to further improve the quality of generations.
G3DR offers diverse and efficient 3D asset generation based on class or text conditioning. 
Despite its simplicity, G3DR is able to beat state-of-the-art methods, improving over them by up to $22\%$ in perceptual metrics and $90\%$ in geometry scores, while needing only half of the training time.
Code is available at \href{https://github.com/preddy5/G3DR}{https://github.com/preddy5/G3DR}.
\end{abstract}

\makeatletter{\renewcommand*{\@makefnmark}{}
\footnotetext{* Authors contributed equally.}\makeatother}

\section{Introduction}
\label{sec:intro}

Generating 3D assets is becoming increasingly vital for applications such as VR/AR, film production, and video games.
Traditionally, 3D modeling is done by devoted artists and content creators, however, nowadays it is desirable to use machine learning solutions to automate the process. 
The seminal work of NeRF \cite{DBLP:conf/eccv/MildenhallSTBRN20} made a large step in 3D novel view synthesis, by posing the problem as learning an implicit radiance field solely from calibrated images.
%
However, the method requires many different views of the same object/scene with known camera transformations to perform an accurate reconstruction.
Furthermore, NeRF can only reconstruct a scene but not generate plausible similar-looking scenes.

Several methods \cite{DBLP:conf/cvpr/ChanMK0W21,DBLP:conf/nips/SchwarzLN020,DBLP:conf/cvpr/Niemeyer021} modified NeRF to have generative capabilities.
While these methods show impressive 3D generative capabilities, they primarily cater to well-curated and aligned datasets featuring similar object categories and structures, necessitating domain-specific 3D knowledge of the category at hand.
This knowledge enables the inference of underlying 3D key points, camera transformation, object scale, and precise cropping.
However, this knowledge required for alignment becomes impractical for extensive multi-category datasets such as ImageNet.

\begin{figure}
    \centering
    \includegraphics[width=\linewidth]{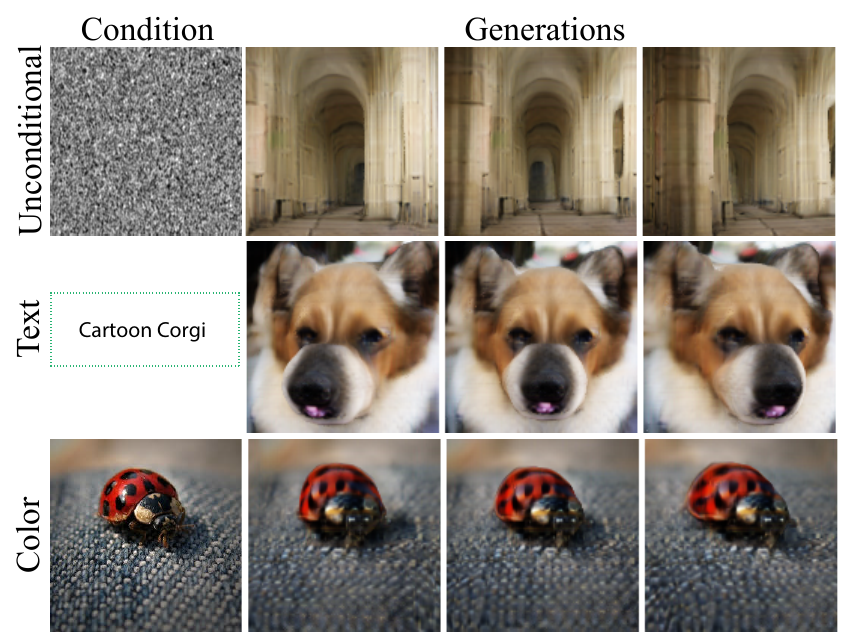}
    \captionof{figure}{Our method is able to generate 3D images conditioned on latents (such as class), text, or images. All the training has been done in ImageNet dataset, that contain only single-view images. 
    }
    \label{fig:teaser}
\end{figure}

In this work, we propose a novel Generative 3D  Reconstruction (G3DR) method that learns from a diverse unaligned 2D dataset such as ImageNet\cite{DBLP:conf/cvpr/DengDSLL009}. 
We combine a latent diffusion model along with a conditional triplane generator to generate highly detailed 3D scenes.  
Training a triplane generator to reconstruct the scene in the input view is a trivial task, and can be done by simply enforcing a reconstruction loss between it and the ground truth.
However, training the model to generate plausible novel views is much more challenging since the ImageNet dataset contains only a \textit{single} image for each scene.
We tackle this by employing a pre-trained language-vision model \cite{DBLP:conf/icml/RadfordKHRGASAM21} to give the necessary supervision for the novel views.

The framework mentioned so far only considers the visual quality of the images and does not ensure plausible geometry.
We use a generic off-the-shelf monocular depth estimation model\cite{miangoleh2021boosting} to estimate the depth map of images in the dataset. 
These depth maps are used to supervise the geometry of the input view.
We do so using a depth reconstruction loss that acts as a surrogate loss for the geometry.
While this loss gives some concept of geometry, in the absence of multi-view data or alignment, optimizing the triplane generator is an overparametrized problem and, thus prone to many solutions, most of which do not ensure good novel views or plausible geometry. 
We ensure that the geometry is faithful to the supervision depth map, we propose a novel depth regularization method.
This depth regularization method scales the gradients of density and color variables of NeRF volumetric rendering using a kernel.
The kernel takes into account the proximity of 3D points that correspond to the density and color value from the camera and the surface's distance from the camera and scales the gradients to encourage high-density values close to the surface.
We further improve the quality of the textures of generated 3D scenes using a multi-resolution triplane sampling strategy.
Our multi-resolution sampling strategy helps in improving the model performance without increasing the number of model weights.

In summary, we make the following \textbf{contributions}:

\begin{itemize}
    \item We design a framework for 3D content generation from a single view. Our method can be coupled with generative diffusion models for unconditional, class-conditional, and text-conditional 3D generation.
    \item We propose a new gradient regularization method in order to preserve the geometry of the objects.
    \item We propose an efficient multi-resolution sampling strategy to enhance the quality of generated images.
    \item We improve the state-of-the-art on Imagenet by $22\%$ in quality and $90\%$ in geometry while lowering the computing by $48\%$. We further validate our method on three other datasets: SDIP Dogs \cite{mokady2022self}, SDIP Elephants \cite{mokady2022self} and LSUN Horses \cite{yu2015lsun}.
\end{itemize}

\section{Related Work}
\label{sec:related_work}

\begin{figure*}
    \centering
    \includegraphics[width=\textwidth]{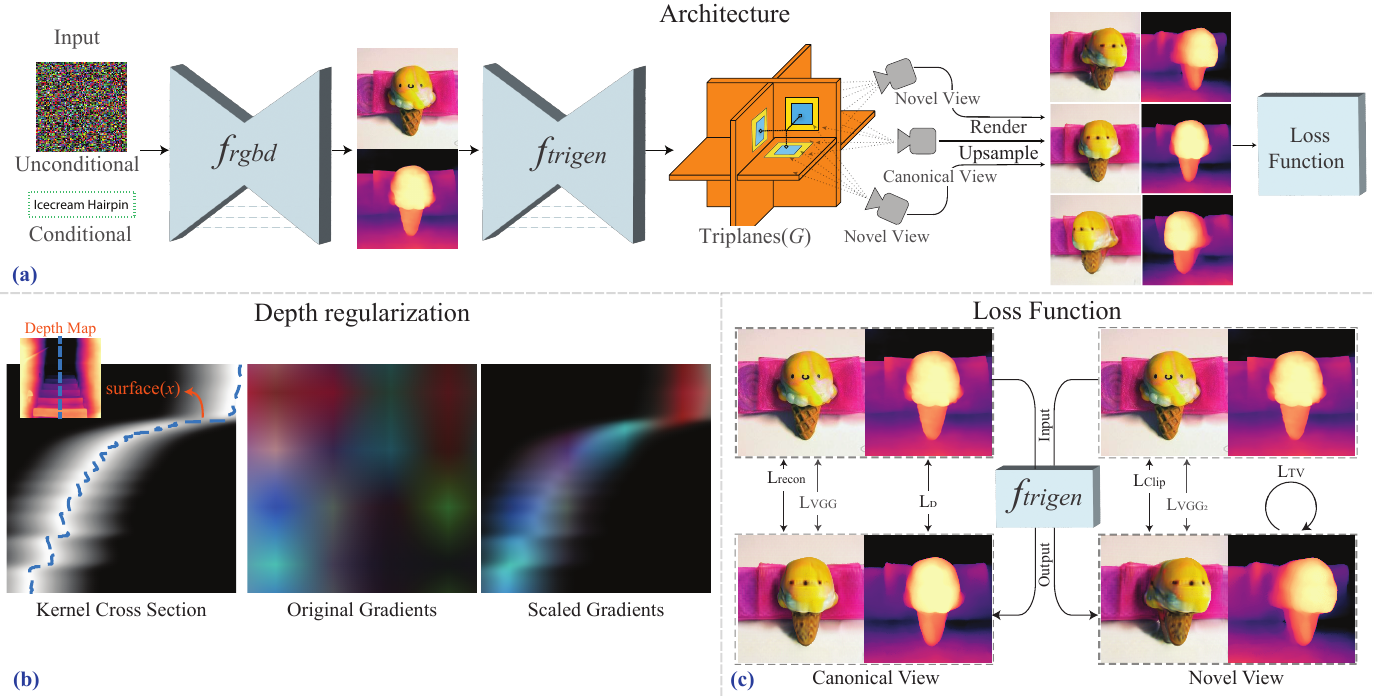}
    \vspace{-7mm}
    \captionof{figure}{a) The architecture of our method. Our framework is conditioned on some visual input, class cateogry or text, and generates an image. Then it feeds that image over a triplane generator, and it finally renders it, ensuring good image quality and geometry using a regularization depth; b) an illustration of our kernel in 2D; the blue line on the \textit{Depth Map} represents the selected cross section, in the \textit{Original Gradients} we visualize high dimensional gradients using rgb channels and \textit{Scaled Gradients} show how the kernel modifies the volume rendering function gradients c) the losses of our model. In the canonical view, our method uses a combination of reconstruction, perceptual and depth loss. In the novel view, it uses a combination of clip, perceptual and tv loss. The losses are scaled accordingly, while the loss gradients during backpropagation are scaled based on the kernel in (b).
    }
    \label{fig:framework}
\end{figure*}

\noindent\textbf{3D aware generation. } Neural Radiance Fields (NeRF) \cite{DBLP:conf/eccv/MildenhallSTBRN20} representation implicitly encodes a scene in the weights of an MLP learning solely from RGB supervision using reconstruction loss.
Many works combined NeRF representation with generative models such as GANs \cite{DBLP:conf/nips/GoodfellowPMXWOCB14} or diffusion processes \cite{DBLP:conf/nips/HoJA20, DBLP:conf/cvpr/RombachBLEO22}.
Some examples of these works include PI-GAN \cite{DBLP:conf/cvpr/ChanMK0W21}, StyleNerf \cite{DBLP:conf/iclr/GuL0T22}, GRAF \cite{DBLP:conf/nips/SchwarzLN020}, GIRAFFE \cite{DBLP:conf/cvpr/Niemeyer021}, and other follow-ups \cite{DBLP:conf/cvpr/XueLSL22, DBLP:conf/cvpr/DengYX022, DBLP:conf/cvpr/CaiODG22, DBLP:conf/cvpr/Tewari0PFAT22, DBLP:conf/nips/SkorokhodovT0W22, DBLP:conf/nips/SchwarzSNL022, DBLP:journals/corr/abs-2110-09788, DBLP:conf/cvpr/WangCH0022, DBLP:conf/cvpr/Or-ElLSSPK22, DBLP:conf/cvpr/ZhangZGZPY22, DBLP:journals/ijcv/ZhangZGZYC23, DBLP:conf/cvpr/XuPYSZ22, DBLP:conf/iclr/PooleJBM23, nguyen2019hologan, Anciukevicius_2023_CVPR, karnewar2023holodiffusion, erkocc2023hyperdiffusion, control3diff, shue20233d}.
While these works show spectacular results, most of them are constrained to work with scenes that contain multiple views or do not scale well to large datasets like shown in  \cite{brock2018large, lee2022autoregressive}. 
Furthermore, using an MLP representation comes with a high training cost and is very GPU-hungry.
EG3D \cite{DBLP:conf/cvpr/ChanLCNPMGGTKKW22} proposes a triplane representation which
scales well with resolution, allowing better generation details and also lowering the training cost.
While we use a similar triplane representation as EG3D, unlike them, we are not constrained to needing the camera transformation for each input image.

\noindent\textbf{Optimization-based methods. }
Works like DietNerf \cite{Jain_2021_ICCV}, Dream Fields \cite{DBLP:conf/cvpr/JainMBAP22} and DreamFusion \cite{DBLP:conf/iclr/PooleJBM23} train a NeRF representation of a scene.
They do so by optimizing an individual NeRF model per-scene conditioned on an input text, something that would be very challenging and expensive for datasets with hundreds of thousands of scenes, e.g., ImageNet.
In contrast, we train an amortized model capable of conditional and unconditional generation.
Furthermore, unlike them, we train a single model for the entire dataset.

\noindent\textbf{Single-view 3D reconstruction in large datasets. } NeRF-VAE \cite{DBLP:conf/icml/KosiorekSZMSMR21} combines a variational autoencoder with NeRF using amortized inference to reconstruct 3D scenes from single-views.
However, it relies on multi-view images during training and uses simple datasets.
LoloNeRF \cite{DBLP:conf/cvpr/RebainMYLT22} learns a generative model of
3D face images achieving good quality in 3D face reconstruction but requires a pretrained keypoint estimator and an optimization of samples outside of the training set.
%
%
3DGP \cite{DBLP:conf/iclr/SkorokhodovSXRL23} and follow ups \cite{DBLP:journals/corr/abs-2302-06833, DBLP:journals/corr/abs-2303-17905} showed that it is possible to do 3D reconstruction of a large single-view dataset that contains $1,000$ classes \cite{DBLP:conf/cvpr/DengDSLL009}.
3DGP based on GANs, designed a framework that is able to generate realistically looking 3D objects from the ImageNet dataset.
To do so, they used a depth estimator for geometry preservation and combined it with a flexible camera model, and a knowledge distillation module.
The method showed promising results, but at the same time leaves room for improvement.
The geometry of the reconstructed images is far from perfect and the visual quality of the generated images contains many artefacts.
Finally, it comes with a massive computational cost.
Our work is motivated by \cite{DBLP:conf/iclr/SkorokhodovSXRL23}, and we try to solve the same problem as them. 
At the same time, we design our method to leverage a novel depth regularization module that improves both the visual quality and the geometry of the generated images, without any need for adversarial training or knowledge distillation.

\noindent\textbf{Depth guidance. } 
Works such as \cite{roessle2022dense, uy2023scade, song2023darf}, have used depth information to enhance NeRF representation novel view reconstruction in the context of capturing a single scene.
GSN \cite{DBLP:conf/iccv/DeVries0STS21} use depth prior for their 3D generation.
Yet, the method is limited to needing ground truth depth, which is not given in most large datasets.
Some other works \cite{DBLP:conf/eccv/ShiSZYC22, DBLP:conf/nips/YuPNS022} bypass this problem by using some depth estimator to generate the depth.
However, the task of these works is limited to geometry reconstruction.
3DGP, like us, uses an off-the-shelf depth estimator to control the geometry.
Their core characteristic is to train a depth adaptor module that mitigates the estimator's depth precision.
At contrast, we design a novel depth regularization module that allows us to do good 3D reconstruction from single-view images.

\section{Methodology}
\label{sec:methodology_alternative}

We describe our problem formulation in Section \ref{sec:problem}. 
We observe that naively training the network is an overparametrized problem that leads to degenerate solutions. Thus, to solve this, in Section \ref{sec:depth_regularization}, we propose a novel depth regularization module.
We describe our multi-resolution sampling in Section \ref{sec:sampling}.
We give our training procedure and losses in Section \ref{sec:training}, and we conclude this part by explaining the generation process in Section \ref{sec:generation}.
We show a visualization of our framework in Fig. \ref{fig:framework}a.

\subsection{Problem formulation}
\label{sec:problem}
Given a latent $c_k$, which could be a class, image, text, or other representation, our goal is to generate a 3D representation $I_k$ by a neural network from the latent $c_k$.
We use a latent diffusion model $f_{rgbd}$ to generate an image $rgbd_k$ with depth.
Then we use a triplane generator $f_{trigen}$ to complete the $rgbd_k$, and subsequently volume render the generated triplanes using $f_{dec}$.

Training $f_{trigen}$ model to generate 3D scenes using an unaligned dataset that contains only a single view per scene remains an ill-posed problem with many possible naive solutions.
This extreme training scenario leads $f_{trigen}$ causes \textit{volume collapse} in the estimate 3D models where the surface is incorrectly modeled using a few disconnected regions of semi-transparent clouds of content which explain the input view but when viewed from another angle results in a blurry or a skewed input image. 
%
%
We solve this through our novel depth regularization approach by adjusting the gradients of NeRF volumetric rendering function during training.
We enhance the texture quality of generated 3D scenes by employing a multi-resolution triplane sampling strategy, improving the model performance without increasing the number of model weights.

\subsection{Depth regularization}
\label{sec:depth_regularization}

%
%
%
%
%
We propose a novel depth regularization technique for efficient training of $f_{trigen}$, enabling the generation of scenes with high-fidelity geometry while preventing volume collapse.
%
Our proposed depth regularization technique is theoretically compatible with most NeRF implementations and does not induce any significant overhead while training. 

%
Let $o$ and $d$ represent the ray origin and ray direction and let $t$ be the sampled distance along the ray.
We use volumetric rendering to render a triplane scene $G$ along the rays $r(t) = o + td$.
We sample the latent values $g$ from axis-aligned orthogonal feature planes of $G$ by projecting $r(t)$ onto the feature planes\cite{DBLP:conf/cvpr/ChanLCNPMGGTKKW22}. 
Then, we use an implicit function $f_{dec}$ to estimate color $c$ and density $\sigma$ conditioned on $g$ and $d$.
Using these $c$ and $\sigma$ values, we approximate the volume rendering integral along the ray:

%
%

\begin{equation}
  \begin{array}{l}
    \mathlarger{w_i = -(1 - exp(-\sigma_i \delta_i)) \ exp\Biggl(\sum\limits_{j=1}^{i-1} \sigma_j} \delta_j \Biggl) \\
    \mathlarger{C(r) = \sum\limits_{i=1}^N w_i c_i} \\
  \end{array}
\end{equation}

\noindent where  $\delta_i =  t_{i+1} - t_i$. 
%
Ideally, $\sigma$ values are expected to be high around the ground-truth surface i.e. if the surface is at a distance $x$ from the ray origin then $\sigma$ is expected to be high where $x \approx t_i$.
However, while using a perspective camera projection-based ray casting setup without any sufficient multi-view supervision, $r(t)$ closer to the camera (i.e., lower values) receive higher gradients than the far points resulting in artifacts and undesired geometry\cite{grad_scaling_euro}. 
To solve this, we use our regularization to encourage high $\sigma$ values closer to the expected surface while discouraging the $\sigma$ values away from the surface.
%
%
Our proposed depth regularization does this by re-scaling the gradients of density and color values w.r.t to the loss function based on the distance between $r(t)$ and the surface using the equation:

\begin{equation}
  \begin{array}{l}
    \quad \mathlarger{\frac{\partial \sigma_i}{\partial \theta} = k(x, t_i) \frac{\partial \sigma_i}{\partial \theta};\quad  \;\;\;
    \frac{\partial c_i}{\partial \theta} = k(x, t_i) \frac{\partial c_i}{\partial \theta}}.\\
  \end{array}
  \label{eq:grad_kernel}
\end{equation}

\noindent We define $k(x, t_i)$ as a kernel:

\begin{equation}
    k(x, t_i)=min(c_{max}, max(s_1 exp\Biggl(-\frac{(x-t_i)^2}{s_2}\Biggl), c_{min}))
  \label{eq:grad_kernel_eq}    
\end{equation}

\noindent where $c_{min}$, $c_{max}$, $s_1$, and $s_2$ are hyperparameters, $x$ is the depth of pixel (the distance of the surface from the ray origin). 
The kernel value is high where $t_i$ is close to the surface i.e., where the absolute difference between $t_i$ and $x$ is low and smoothly decreases as the absolute difference increases.
%
The values of $c_{min}$ and $c_{max}$ to determine the maximum and minimum values of the kernel. 
The $c_{min}$ value is required to be a positive non-zero so that densities away from the surface reduce from their default initialization values over the course of the training.
The hyperparameters $s_1$ and $s_2$ control the spread of the regularization kernel around the surface.
A very high $s_2$ nullifies the effect of regularization and a very low $s_2$ results in vanishing gradients. 
We have empirically observed choosing $s_2$ value based on the ray sampling density gave the best results, and heuristically set it to half of the distance between the coarse ray samples.
%
%
%
In Fig \ref{fig:framework}b we show an illustration of how the gradients are scaled based on a 2D cross-section of a depth map. 

\subsection{Multi-resolution sampling}
\label{sec:sampling}
\begin{figure}[t!]
    \centering
    \includegraphics[width=\columnwidth]{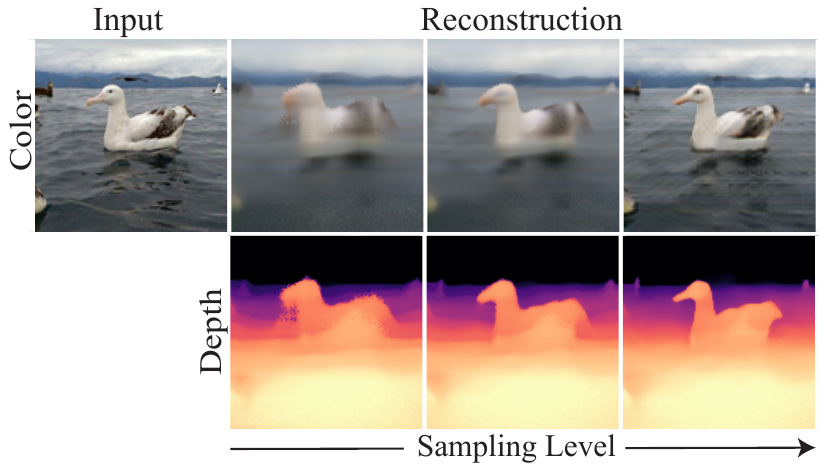}
    \caption{
    An illustration of our multi-resolution sampling. We observe how increasing the sampling level directly increases the reconstruction quality.
    }
    \label{fig:lod}
\end{figure}

We adopt a multi-resolution triplane sampling strategy to improve the generation model performance.
Given a triplane $G$ we create a set of tri-planes with different resolutions $\{G_l\}_{l=1}^L$, where L represents the total number of levels. 
Each level $G_l$ is constructed by resampling the previous level $G_{l-1}$ to half the resolution.
In our experiments, we construct 3 levels. 
All levels are resampled with antialiasing to minimize undesirable distortion artifacts. 
For each $G_l$, we sample latent values corresponding to $r(t)$ by projecting $r(t)$ onto each of the orthogonal feature planes and retrieving their corresponding feature vectors using bilinear interpolation. 
We then aggregate feature vectors from individual planes using 
 mean operation and we collate feature vectors from different $G_l$ using summation \cite{DBLP:conf/cvpr/ChanLCNPMGGTKKW22, takikawa2021nglod}. 
%
%
Note that our sampling strategy is unlike the \textit{multi-resolution triplanes strategy} used in methods like \cite{kplanes_2023, zhuang2023anti,DBLP:conf/iccv/HuWMY0LM23} where separate triplanes latents are learned at different resolutions.
This style of multi-resolution sampling improves model performance without any increase in the number of model parameters.
We then pass the final feature vectors to $f_{dec}$ to generate the 3D features of the image, which are then rendered via neural volume rendering. 
In Fig.~\ref{fig:lod} we show how texture and geometry are affected if we start sampling only from the coarsest level and then add samples from finer levels.

\subsection{Training}
\label{sec:training}

In this section, we describe the different losses used in our framework. We distinguish between the training for the canonical view and the one for novel views.

\textbf{Canonical view. }
Along the canonical view, we train the network to accurately reconstruct the ground truth with good geometry.
The ground truth camera extrinsic parameters for all the datasets in our experiments are unknown and there is no accurate model to estimate these parameters.
Thus, we choose a reasonable set of camera parameters and use them as the canonical view parameters for all the images.
We follow the standard formulation, and define the reconstruction loss $L_{recon}$
as the total squared error between the rendered and the true pixel values. 

%

We also use a depth loss $L_D$ defined as $L_1$ loss between the pseudo-ground truth depth map, and we estimate the accumulated depth values of the rendered images:
\begin{equation}
D = \sum\limits_{i=1}^N t_i w_i/\sum\limits_{i=1}^N w_i +\epsilon. 
\label{eq:acc_depth}
\end{equation}

\noindent where $\epsilon$ is a hyperparameter introduced for training stability. This approximation of depth can result in the same $D$ value for different $\{w_i\}_{i=1}^N$ values many of which are degenerate causing volume collapse. We resolve this by manipulating the gradients as explained in Section \ref{sec:depth_regularization} during backpropagation.

To further improve the visual performance, we add perceptual loss $L_{VGG}$ \cite{zhang2018unreasonable}. We define our loss for the canonical view as a weighted sum of the mentioned losses:

\begin{equation}
    L_{canon} = \lambda_1 L_{recon} + \lambda_2 L_D + \lambda_3 L_{VGG}
\end{equation}

\noindent with $\lambda_1, \lambda_2, \lambda_3$ being hyperparameters that scale the losses.

\textbf{Novel view. } A main challenge of generating novel views from single views is the loss supervision in novel views. Other works do this either by using adversarial training \cite{DBLP:conf/iclr/SkorokhodovSXRL23,DBLP:journals/corr/abs-2302-06833} or 3D-aware inpainting \cite{DBLP:journals/corr/abs-2303-17905}.
Instead, we design our novel framework to use a loss $L_{CLIP}$ based on the difference of features between the novel views and ground truth, using a visual-language model  \cite{DBLP:conf/icml/RadfordKHRGASAM21}.
Our intuition is that despite the camera movements, the semantics of an image should be the same as that of the ground-truth image.
We show that this solution, despite being very simple, is powerful enough, and by using it, the network gets the needed supervision for the novel views.
Furthermore, because it has no adversarial training, its convergence is relatively more stable than the other methods.
For geometry supervision, we use the TV-loss ($L_{TV})$ \cite{DBLP:conf/cvpr/NiemeyerBMS0R22} over the accumulated depth to encourage smooth geometry, while for photorealism we use a perceptual loss. We define our loss for novel views as a weighted sum of the mentioned losses:

\begin{equation}
    L_{novel} = \lambda_4 L_{CLIP} + \lambda_5 L_{TV} + \lambda_6 L_{VGG_2}.
\end{equation}

\noindent For $L_{VGG_2}$, unlike in canonical view where we use features from five levels, in novel view, we use features only from the last two levels. 
This is because while we expect the semantics of the image in the novel view to be the same as the input image, its low-level features are not necessarily the same, thus using all five features results in blurry images.
To compensate for using fewer features, we set $\lambda_6$ to twice the value of $\lambda_3$.

\textbf{Alternating between novel and canonical view. } We randomly sample between the canonical and novel views during training.
We design a heuristic probabilistic sampling, where we initially sample with higher probability for the canonical view.
In this way, the network quickly learns the easier task of image reconstruction.
We linearly increase the probability of sampling for the novel views during the training, but the probability of sampling a novel view never gets higher than that of sampling the canonical view. 
We show a visualization of our losses in Fig. \ref{fig:framework}c and give more details in the supplementary material.

\subsection{Generation}
\label{sec:generation}

Our network described so far, would be powerful to do 3D generative reconstruction of $rgbd$ images, a task which might be important for AR/VR. 
However, in this work, we focus on unconditional or class-conditioned generative modeling, to diverse generate 3D scenes.
Thus, we first train a diffusion model in ImageNet that is able to generate realistically single-view $rgbd$ images. 
%

We use our diffusion model for unconditional or class-conditioned generation of $rgbd$ images.
We then feed the generated image to our model as trained in Section \ref{sec:training} which completes the $rgbd$ images by generating a 3D triplane.
Interestingly, in our experimental section, we show that our model not only works for ImageNet-like images conditioned on classes but also on images conditioned on other modalities, such as text. To do so, we sample from a text-to-image latent diffusion model, and then feed those images into our model to get their 3D representation.
Perhaps surprisingly, we show that our model works well even for out-of-domain samples, such as cartoons.

Finally, it is expensive to train a model that produces high-resolution 3D images.
Instead, we train our model to generate moderate-resolution images (e.g., 128x128).
Then, we use a super-resolution network \cite{DBLP:conf/eccv/ZhangLLWZF18}, to upsample the images to the desirable resolution (e.g., 256x256). 
We perform this upsampling during both the training and sampling procedures.
\section{Experiments}
\label{sec:experiments}

\begin{figure*}[!t]
    \centering
    \includegraphics[width=\textwidth]{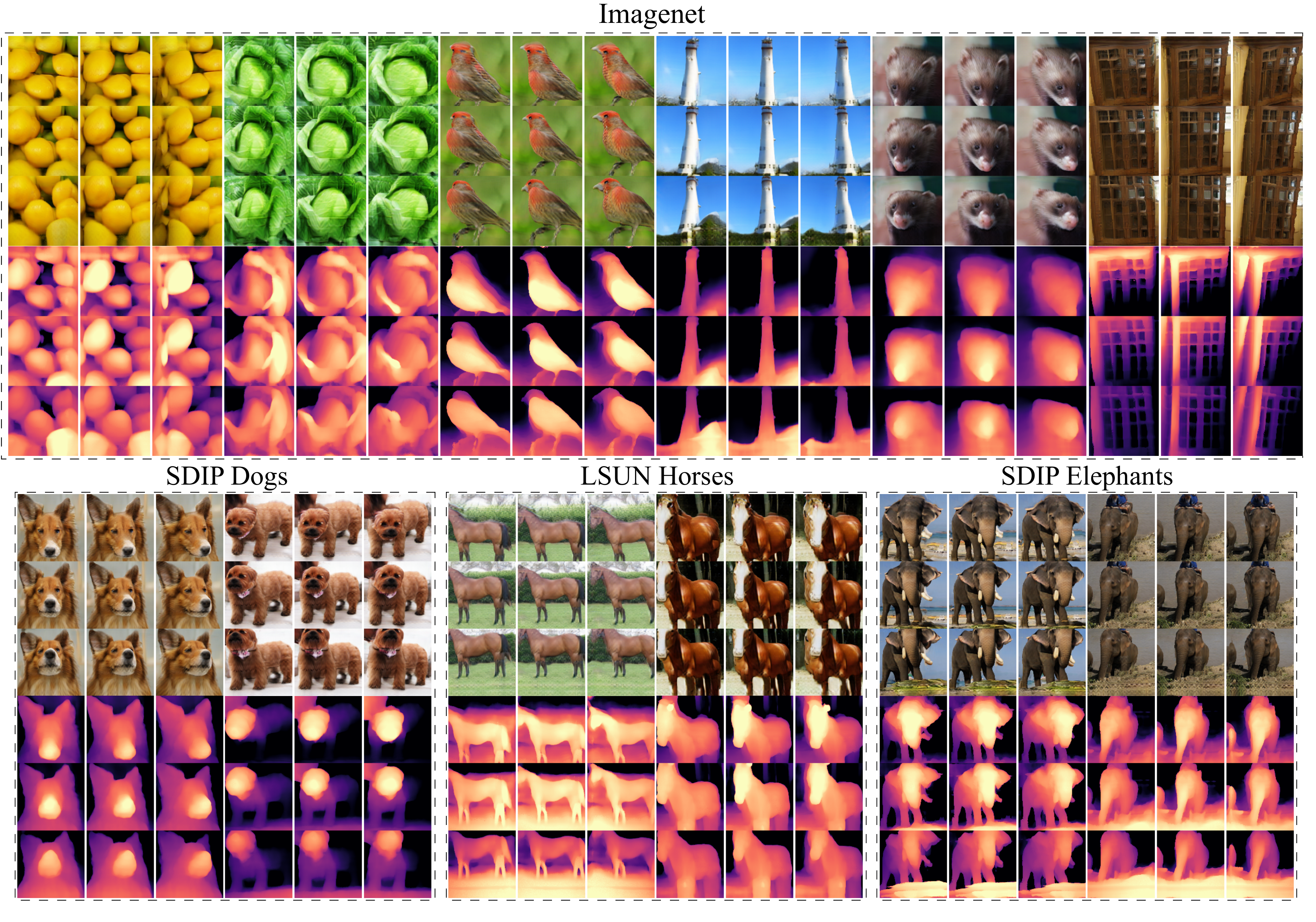}
    \vspace{-7mm}
    \captionof{figure}{Qualitative results of our method. In the first row we present qualitative results generated in ImageNet dataset and their corresponding depth. In the second row we show qualitative results in the finegrained datasets with their corresponding depth.
    }
    \label{fig:all_results}
\end{figure*}

\subsection{Experimental setup}

\noindent \textbf{Implementation details. }
We generate $rgbd_k$ conditioned on $c_k$ using $f_{rgbd}$ a latent diffusion model \cite{stan2023ldm3d}. We then feed the images to $f_{trigen}$ a U-net containing x residual blocks, to generate triplanes consisting of 64 channels. We render the triplanes using small MLPs with 2 layers and 64 hidden units each with a Softplus activation in between.
Finally, we feed the rendered images into a super-resolution network \cite{DBLP:conf/eccv/ZhangLLWZF18}.
We train our model for $400,000$ iterations with batch size $64$, using Adam optimizer \cite{DBLP:journals/corr/KingmaB14} with initial learning rate of $0.001$.
We lower the learning rate using a cosine annealing strategy \cite{DBLP:conf/iclr/LoshchilovH17}.
During training, we alternate between optimizing for the canonical and novel view, based on a probabilistic procedure as explained in the supplementary.
We balance the losses by setting weights of reconstruction, depth, $vgg$, clip, tv and $vgg_2$ loss ($\lambda_1, \lambda_2, \lambda_3, \lambda_4, \lambda_5, \lambda_6$) to $1$, $2$, $0.5$, $0.35$, $0.1$, $0.5$.
For novel views camera parameters, we generate rays by sampling for yaw from $\mathcal{N}(\pi/2, 0.3)$ and for pitch from $\mathcal{N}(0, 0.15)$.

\noindent \textbf{Datasets. } We use $4$ standard datasets for our experiments: ImageNet \cite{DBLP:conf/cvpr/DengDSLL009}, Dogs \cite{DBLP:conf/siggraph/MokadyTYLMDCI22}, SDIP elephants \cite{DBLP:conf/siggraph/MokadyTYLMDCI22}  and LSUN Horses \cite{DBLP:journals/corr/YuZSSX15}.
We perform our main experiments in ImageNet, a realistic multi-category dataset containing over one million images divided into $1,000$ classes.
For fair comparisons, we follow \cite{DBLP:conf/iclr/SkorokhodovSXRL23} and filter our $2/3$ of the images.
We do complementary experiments in the other three datasets, following \cite{DBLP:conf/iclr/SkorokhodovSXRL23} to remove the outliers from SDIP Dogs and LSUN Horses reducing their sizes to $40,000$ samples.

\noindent \textbf{Metrics. }
We use FID \cite{DBLP:conf/nips/HeuselRUNH17} and Inception Score (IS) \cite{DBLP:conf/nips/SalimansGZCRCC16} to measure the image quality.
They have been originally developed to evaluate the quality of images produced by GANs, but are widely used to measure the image quality in all problems.
While our networks have been trained in a filtered version of ImageNet, we compute the metrics in the full ImageNet.
There are no established protocols to measure the geometry quality of 3D generators.
Some papers use Non-Flatness Score (NFS) \cite{DBLP:conf/iclr/SkorokhodovSXRL23} computed as the average entropy of the normalized depth maps histograms, while others use the depth accuracy, computed as the normalized L2 score between the predicted depth and the pseudo-ground-truth depth.
We compare with other works in both scores.

\begin{table}[!b]
\centering
\resizebox{0.48\textwidth}{!}{
\begin{tabular}{@{}ll|cc@{}}
\toprule
Method                            & Synthesis  & FID $\downarrow$    & IS $\uparrow$   \\ \hline
BigGAN \cite{DBLP:conf/iclr/BrockDS19} \small{ArXiV18}              & 2D             & 8.7    & 142.3   \\
StyleGAN-XL \cite{DBLP:conf/siggraph/SauerS022} \small{SIGGRAPH22}       & 2D             & 2.3    & 265.1   \\
ADM \cite{DBLP:conf/nips/DhariwalN21} \small{NeurIPS21}            & 2D             & 4.6   & 186.7  \\ \hline
IVID 128x \cite{DBLP:journals/corr/abs-2303-17905} \small{ICCV23}            & 2.5D             & 14.1   & 61.4      \\ 
Ours  128x                        & 3D             & \textbf{13.0}    &  \textbf{136.4}   \\ \hline \hline
EG3D \cite{DBLP:conf/cvpr/ChanLCNPMGGTKKW22} \small{CVPR22}                  & 3D-a       & 25.6   & 57.3    \\
StyleNeRF \cite{DBLP:conf/iclr/GuL0T22} \small{ICML22}             & 3D-a       & 56.5  & 21.8   \\
3DPhoto \cite{DBLP:conf/cvpr/ShihSKH20} \small{CVPR20} & 3D-a       & 116.6 & 9.5   \\ 
EpiGRAF \cite{DBLP:conf/nips/SkorokhodovT0W22} \small{NeurIPS22}            & 3D             & 58.2  & 20.4  \\
3DGP \cite{DBLP:conf/iclr/SkorokhodovSXRL23}   \small{ICLR23}               & 3D             & 19.7  & 124.8   \\
VQ3D \cite{DBLP:journals/corr/abs-2302-06833}   \small{ICCV23}               & 3D             & 16.8   &   n/a      \\ \hline
Ours                              & 3D             & \textbf{13.1}    &  \textbf{151.7}     \\ 
\bottomrule 
\end{tabular}}
\caption{Comparison between different generators on ImageNet $256^2$. 3D-a means 3D-aware, 2.5D means autoregressive 2D model that gives emergence 3D properties.}
\label{res:imagenet}
\end{table}

\begin{table}[!t]
\centering
\resizebox{0.4\textwidth}{!}{
\begin{tabular}{lll} 
\toprule
Method                           & Depth accuracy $\downarrow$   & NFS $\uparrow$    \\ \hline
3DGP  \cite{DBLP:conf/iclr/SkorokhodovSXRL23}   \small{ICLR23}           & 0.47   & 18.5         \\ 
IVID  \cite{DBLP:journals/corr/abs-2303-17905}  \small{ICCV23}           & 1.23   & 19.2    \\ \hline
Ours                  & \textbf{0.39}    &  \textbf{36.5}    \\ 
\bottomrule 
\end{tabular}
}
\caption{Geometry comparison between our method and two state-of-the-art methods. We compare in depth accuracy and the NFS metric, reaching better results in both of them.}
\label{res:geometry}
\end{table}
\begin{table}[t!]
\resizebox{0.47\textwidth}{!}{
\centering
\begin{tabular}{lllllll}
\toprule
  &  \multicolumn{2}{c}{Dogs} & \multicolumn{2}{c}{Horses} & \multicolumn{2}{c}{Elephants}\\
Method           & FID$\downarrow$ & NFS $\uparrow$   & FID$\downarrow$ &NFS $\uparrow$ & FID$\downarrow$ & NFS $\uparrow$\\ \hline
Eg3D             & 9.83 & 11.91 & \textbf{2.61} & 13.34& \textbf{3.15}    &2.59        \\ 
EpiGRAF          & 17.3 & 3.53  & 5.82 &  9.73& 7.25    &12.9         \\ 
IVID             & 14.7 & N/A   & 10.2 &   N/A& 11.0    &N/A         \\ 
3DGP             & 8.74 & 34.35 & 4.86 &  30.4& 5.79    &32.8         \\  \hline
Ours             & \textbf{8.37} & \textbf{36.89}  & 5.64 &  \textbf{36.2} & 5.30  & \textbf{35.6}       \\ 
\bottomrule 
\end{tabular}
}
\caption{Results on finegrained datasets. Our method reaches competitive results in quality and the best results by far in geometry.}
\label{res:ablation}
\end{table}

\begin{figure*}[!t]
    \centering
    \includegraphics[width=\textwidth]{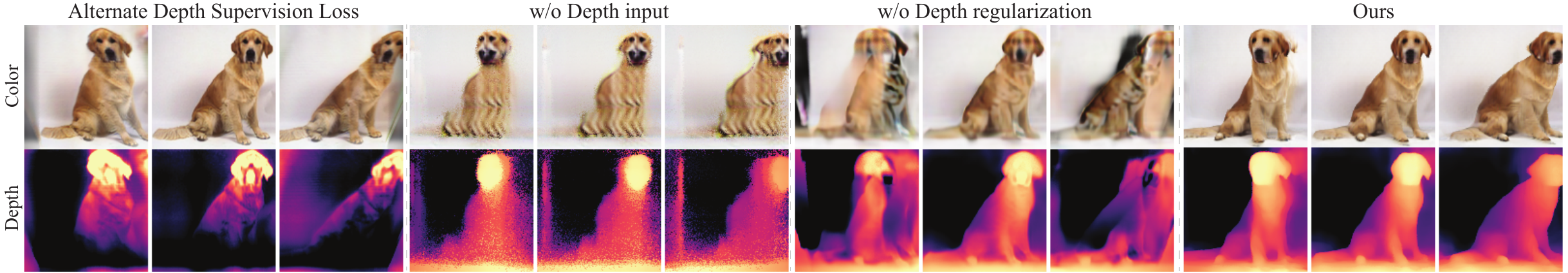} 
    \vspace{-7mm}
    \captionof{figure}{Qualitative evaluations of our method and alternative depth supervision methods.
    }
    \label{fig:ablation_ds_dis_ours}
\end{figure*}%

\subsection{Results}

\textbf{Results on ImageNet. } We present our results in the ImageNet dataset in Tab. \ref{res:imagenet}.
As shown, our method significantly outperforms the other 3D methods.
EG3D which can be considered as the baseline 3D method has an FID score of $25.6$. Other methods improve over it, with the  recent methods, 3DGP and VQ3D reaching FID scores of $19.7$ and $16.8$  
Our G3DR improves the FID score, reaching $13.1$, a relative improvement of over $22\%$, and setting a new state-of-the-art. 
Note that IVID does the evaluation in 128x, reaching $14.1$, which is $1.1$ percentage points ($pp$) worse than our results in that resolution. 
%
%
Similarly, we considerably improve the state-of-the-art in Inception score, reaching $151.7$, a relative improvement of $21.5\%$ over 3DGP.
2D methods, such as BigGAN, StyleGAN-XL, or ADM reach better visual quality scores, but they do not have any consideration for the geometry of the images and, thus are not comparable with our method.

We also evaluate the geometry of the images generated by our method and compare it with the results of 3DGP and IVID. We present the results in Tab. \ref{res:geometry}. 
They reach NFS scores of $18.5$ respectively $19.2$, while our G3DR method reaches an NFS score or $36.5$, almost doubling over the state-of-the-art.
Furthermore, when evaluating the depth accuracy, we outperform all the other methods.
G3DR reaches a depth accuracy of $0.39$, much better than 3DGP and IVID that reach depth accuracies of $0.47$ and $1.33$.

\begin{figure}[t!]
    \centering
    \includegraphics[width=\columnwidth]{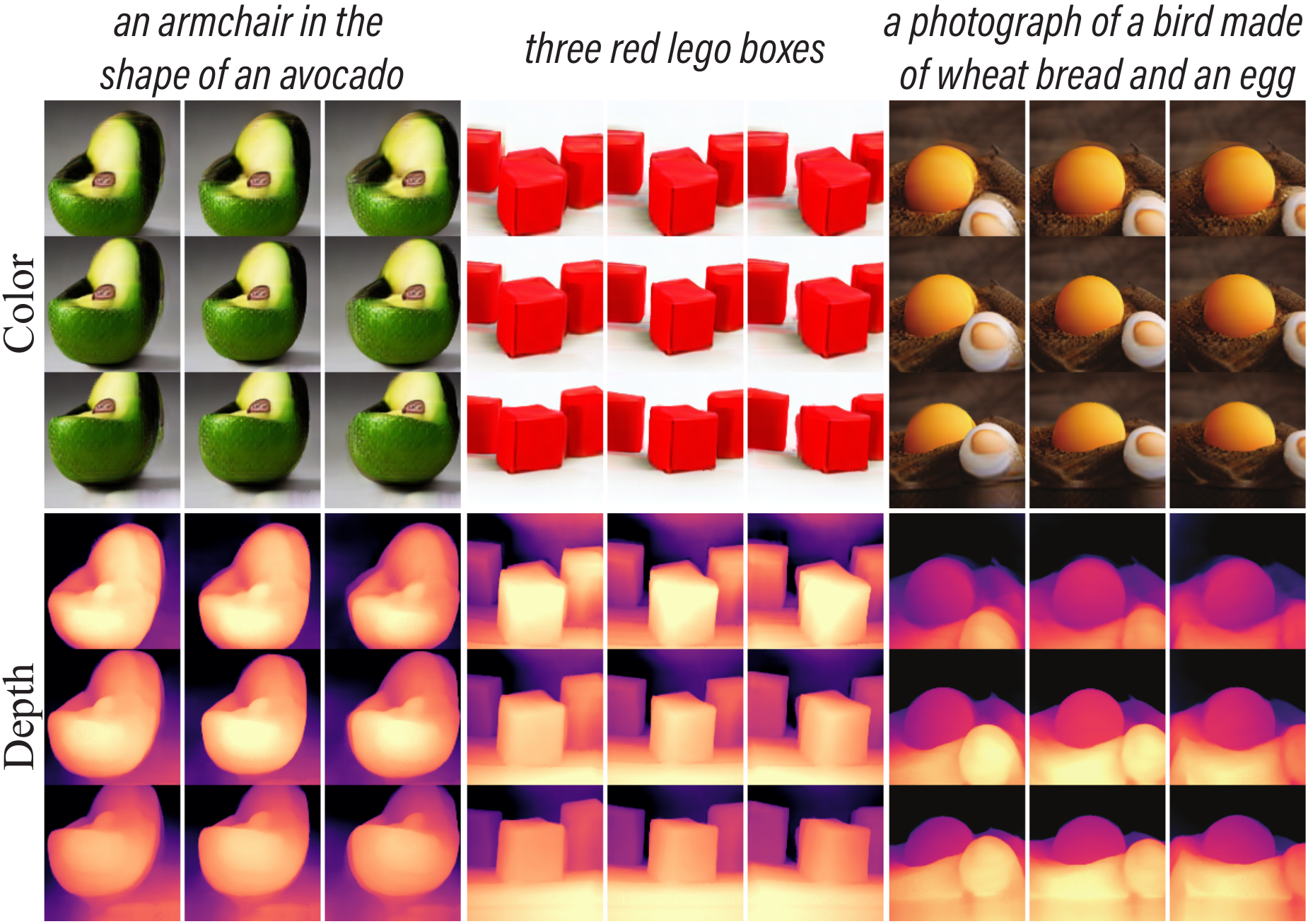}
    \caption{Results on text to 3D, including completely out of domain examples (middle figure).
    }
    \label{fig:ood}
\end{figure}

\textbf{Results on finegrained datasets. } We present the results of our method, and compare it with the competing methods in 3 finegrained datasets: Dogs \cite{DBLP:conf/siggraph/MokadyTYLMDCI22},  LSUN Horses \cite{DBLP:journals/corr/YuZSSX15} and SDIP elephants \cite{DBLP:conf/siggraph/MokadyTYLMDCI22}.
In Dogs dataset, G3DR reaches $8.37$ FID and $36.9$ NFS scores, in both cases surpassing the previous state-of-the-art methods. 
In LSUN Horses, G3DR reaches $5.64$ FID score. While it is not as good as some of the other methods, this is because the other methods sacrifice the geometry of the generated images. For example, EG3D which reaches the best FID score of $2.61$, reaches only $13.34$ NFS, less than half of our score. 
Similarly, in SDIP elephants, EG3D reaches a better FID than our G3DR, but at the cost of more than $10$ times lower NFS.
Overall, G3DR reaches competitive visual quality, while having by far the best geometry of all the methods.

\textbf{Qualitative results. } In Fig. \ref{fig:all_results} we show an extensive qualitative evaluation of G3DR in all four datasets. We observe that G3DR generates high-quality images in novel views while preserving a high level of geometry. We give corresponding videos in the supplementary.

\subsection{Ablation studies}
\label{sec:ablation}
\textbf{Ablation of each block. } We quantify the effect of each of our three main contributions: depth regularization, CLIP supervision and multi-resolution sampling. We do so by training models with the corresponding module turned off.
We observe that turning off the depth regularization has a massive effect on the performance of our method. While rgb generation quality decreases slightly, the geometry quality downgrades massively. 
The FID score downgrades from $13.1$ to $18.1$, the Inception Score decreases from $151.7$ to $116.7$, the NFS lowers from $36.5$ to $25.5$ and the depth accuracy falls from $0.39$ to $1.38$.
This is because our novel regularization method prevents volume collapse that naturally impacts both the quality and the geometry of the generation.
We also see that turning off CLIP comes with a large performance degradation, especially in image quality.
Turning off multi-resolution sampling while affecting the rgb generations does not heavily degrade the geometry.

\textbf{Do we need depth?} We control the geometry of the generated images using depth. We do an experiment proving that using depth is necessary to get good geometry and quality. We present the results in Tab. \ref{res:ablation_depth}, where we show that we reach $64.9$ FID and $30.3$ NFS without a depth map as an input, significantly lower than the results of our G3DR.

\textbf{Alternative depth supervision.} Instead of using our depth regularization module, we use an alternative approach.
More precisely, we integrate the depth loss of Deng et al. \cite{kangle2021dsnerf} in our framework, calling it "Alternative depth".
The method supervises the $w$ values using the KL divergence between the $w$ values and the depth map. 
We present the results in Tab. \ref{res:ablation_depth}, showing that it falls short of our method in both generation quality and geometry.
We give a qualitative example of it and other methods in Fig.  ~\ref{fig:ablation_ds_dis_ours}

\begin{table}[t!]
\centering
\begin{tabular}{lllll} 
\toprule
Method                            & FID $\downarrow$    & IS $\uparrow$    & NFS $\uparrow$  & DA $\downarrow$\\ \hline
w/o Depth regular.            &  18.1 & 116.7& 25.5&     1.38  \\ 
w/o CLIP                   & 19.9  & 100.9 & 35.1 & 0.52\\ 
w/o Multi-res. sampling                  & 14.5  & 128.6 & 36.1 & 0.39\\ \hline
Ours                              & \textbf{13.1}    &  \textbf{151.7}     & \textbf{36.5}  &  \textbf{0.39} \\ 
\bottomrule 
\end{tabular}
\caption{The effect of each block in our framework. We observe that removing each block comes with a decrease in the performance, with the biggest decrease coming if we remove our depth regularization module.}
\label{res:ablation}
\end{table}
\begin{table}[t!]
\centering
\begin{tabular}{lllll} 
\toprule
Method                            & FID $\downarrow$    & IS $\uparrow$    & NFS $\uparrow$  & DA $\downarrow$\\ \hline
w/o Depth input            & 64.9   & 22.2  &     30.3   &   1.20  \\ 
Alternative depth                   & 44.8  &83.0 & 28.9 & 2.07\\ \hline
Ours                              & \textbf{13.1}    &  \textbf{151.7}     & \textbf{36.5}  & \textbf{0.39} \\ 
\bottomrule 
\end{tabular}
\caption{Comparison of our model with methods that do not use depth supervision and that use an alternative depth supervision.}
\label{res:ablation_depth}
\vspace{0.5cm}
\end{table}
\begin{table}[t!]
\centering
\begin{tabular}{llll} 
\toprule
Yaw trans. range                  & FID $\downarrow$    & IS $\uparrow$    & NFS $\uparrow$    \\ \hline
$\mathcal{U}(-00^{\circ},00^{\circ})$             & 11.89   & \textbf{201.56}  &    35.64         \\ 
$\mathcal{U}(-09^{\circ},09^{\circ})$             & \textbf{11.79}   & 193.57  &    35.95         \\ 
$\mathcal{U}(-18^{\circ},18^{\circ})$             & 11.94   & 180.33  &    35.36         \\ 
$\mathcal{U}(-25^{\circ},25^{\circ})$             & 12.39   & 166.74  &    36.42         \\ 
$\mathcal{U}(-35^{\circ},35^{\circ})$             & 13.75   & 145.66  &    \textbf{36.75}         \\ 
$\mathcal{U}(-50^{\circ},50^{\circ})$             & 17.68   & 113.28  &    36.23         \\ 
\bottomrule 
\end{tabular}
\caption{We present the FID, IS, and NFS metrics by uniformly sampling the yaw camera parameter in different ranges.}
\label{res:yaw}
\vspace{0.4cm}
\end{table}

\textbf{Results on different yaw ranges. } We experiment with changing the camera yaw for the novel poses.
Obviously, if we increase the yaw range we expect the generation quality to lower. This is because high movements on camera should result in more extreme novel views than when we have little to no movement in camera (canonical view).
We present the results in Tab. \ref{res:yaw}.
We use $\mathcal{U}(-x^{\circ},x^{\circ})$ to note that the yaw parameter is uniformly sampled in the range [-x, x]. We observe that when the yaw range is low, the quality of generated images barely changes. Sampling yaw from ranges $\mathcal{U}(-09^{\circ},09^{\circ})$ and $\mathcal{U}(-18^{\circ},18^{\circ})$ gives virtually the same FID score as the images in canonical view $\mathcal{U}(-09^{\circ},09^{\circ})$ and slightly worse IS.
After that, increasing the yaw range starts degrading all the scores, with yaw $\mathcal{U}(-50^{\circ},50^{\circ})$ having FID that is $6$ points worse than that of the canonical score.
When it comes to the geometry of the generations in novel view, we observe that the NFS values reach the same value regardless of the yaw transformation.



\textbf{Results on different modalities and domains. } We do an intriguing experiment, checking the effectiveness of our method in a different modality, conditioning our model in text. We show the results in Fig. \ref{fig:ood}.
We observe that our method generates high-quality 3D reconstruction of text prompts, despite it has never been trained on them.
More interestingly, our method shows high results even when used for completely out-of-domain generations, such as cartoons, astronomical concepts, or geometric shapes.
We provide additional images and videos in the supplementary.

\textbf{Convergence. } We use only $400,000$ training steps, much fewer than competing methods such as IVID \cite{DBLP:journals/corr/abs-2303-17905} which uses over a million steps.
Our total training time is $14.5$ A100 days, half as much as 3DGP \cite{DBLP:conf/iclr/SkorokhodovSXRL23} that needs $28$ A100 days.

\section{Conclusion}
\label{sec:conclusion}

In this work, we developed a novel framework for the task of 3D generation on large, multi-category datasets, such as ImageNet.
At the core of our method is a novel depth regularization method that allows our framework to reach very high geometry.
Unlike other methods, that use adversarial training for novel views, we combine our geometry module with a visual-language module for novel-view supervision.
For higher visual quality, we also introduce a new sampling method.
We show in our experimental section, that despite lowering the training time by $48\%$, our method is able to surpass the state-of-the-art by $90\%$ in geometry and $22\%$ in visual quality.




{\small
\bibliographystyle{ieee_fullname}
\bibliography{egbib}
}

\end{document}